\documentclass[10pt,twocolumn,letterpaper]{article}
\usepackage[pagenumbers]{cvpr}
\usepackage{booktabs}
\usepackage{amsmath}
\usepackage{graphicx}
\usepackage{array}
\usepackage{xspace}
\definecolor{cvprblue}{rgb}{0.21,0.49,0.74}
\usepackage[pagebackref,breaklinks,colorlinks,allcolors=cvprblue]{hyperref}

\def\confName{CVPR}
\def\confYear{2026}
\newcommand{\vrrqa}{\texttt{VRR-QA}\xspace}

\title{Question-Aware Evidence Ledgers for Video Relational Reasoning}

\author{Yilin Ou \quad Mengshi Qi\footnote{Corresponding author.} \quad Huadong Ma\\
State Key Laboratory of Networking and Switching Technology\\
Beijing University of Posts and Telecommunications, China\\
{\tt\small qms@bupt.edu.cn}
}

\begin{document}
\maketitle

\begin{abstract}
The \vrrqa challenge evaluates visual relational reasoning in videos, where
answers often depend on implicit spatial relations, event boundaries, target
identity, and dialogue context rather than a single salient frame. We present a
test-time reasoning pipeline built around a strong GPT-5.5 video QA solver and a
set of question-aware evidence ledgers. The initial solver answers each question
from a uniform video representation, while routed ledgers are prompted to make
the required targets, count units, reference frames, and temporal or spatial
scope explicit for counting, spatial, endpoint, viewpoint, and dialogue
reasoning.
External tools such as open-vocabulary detection, depth cues, pair crops, ASR,
and scene-graph ledgers are used only as evidence sources. A conservative gate
keeps the current answer unless independent evidence uniquely supports a
different option. The final evidence-gated pipeline achieves 92.95\%
overall accuracy and 93.79\% macro accuracy on the challenge test split.
\end{abstract}

\section{Introduction}
Recent vision-language and video-language models have made strong progress on
open-ended image and video understanding
\cite{radford2021clip,alayrac2022flamingo,li2023blip2,liu2023llava,maaz2023videochatgpt,lin2023videollava}.
The \vrrqa challenge exposes a different failure mode: many questions require
precise relational evidence. A model may recognize the scene and objects yet
still confuse screen-left with actor-left, count repeated views as new
instances, use an intermediate frame for a final-state question, or infer a
speaker-addressee relation from visual context alone.

These errors often begin before visual recognition: the system must decide what
the question is asking. Words such as ``first'', ``last'', ``front'', ``behind'',
``left'', ``right'', ``start'', and ``end'' can refer to time, image layout,
object-intrinsic axes, a path or queue, a character viewpoint, or an endpoint
state. We therefore avoid training a new model and instead build a test-time
system that audits the answer with question-aware structured evidence.

The core design is a routed ledger. Each question receives an initial answer,
then only the relevant reasoning families are invoked. Counting questions
receive instance and event ledgers; spatial questions receive target-reference
and coordinate-frame ledgers; endpoint questions receive start/end state checks;
dialogue questions receive ASR and visual grounding. The final merge is
deliberately conservative: the system changes an answer only when the evidence
is sufficient and the new option is uniquely supported.

We present the system as a single pipeline with a fixed initial solver, fixed
routers, structured evidence modules, and conservative merge rules. This framing
keeps the method boundary clear: each answer revision must be justified by a
routing rule, a textual pattern, a target family, candidate consensus, or
answer-independent evidence quality before it can be accepted.

Our main contributions are: (1) question-aware evidence ledgers for resolving
count units, target-reference binding, and reference-frame ambiguity, (2) a
routed evidence-ledger formulation for test-time video relational reasoning,
(3) conservative answer revision rules that separate candidate generation from
answer acceptance, and (4) an implementation that combines strong multimodal
prompting with external evidence without training task-specific parameters.

\section{Task}
Video question answering benchmarks have emphasized localized evidence,
temporal actions, and long-form comprehension
\cite{lei2018tvqa,xiao2021nextqa,mangalam2023egoschema}. \vrrqa is a
multiple-choice video question answering benchmark for visual relational
reasoning \cite{vrrqa2025}. Each example contains a video clip, a question, and
up to eight answer options. The required output is a single option letter. The
benchmark spans counting, vertical and lateral spatial relations, relative
depth and proximity, motion, viewpoint, causal reasoning, physical context, and
social interaction.

\section{Related Work and Positioning}
Our setting is connected to several lines of video and multimodal
understanding. Video classification, retrieval, captioning, and temporal action
localization methods have studied how to preserve temporal dynamics, salient
snippets, and interaction structure across frames
\cite{qi2020fewshot,qi2021semanticsaware,qi2020sportscaptioning,yun2024salientsnippet}.
These directions motivate our use of temporal ledgers: for relational video QA,
the relevant evidence is often a bounded event, an endpoint state, or an
interaction trajectory rather than a single global video label.

Fine-grained human-centric understanding further highlights the importance of
pose, motion, and modality-specific evidence. Prior work on action quality and
action form assessment uses pose-guided contrastive regression and multimodal
chain-of-thought explanations to evaluate how actions are performed, not only
which action occurs \cite{qi2025aqa,qi2026afa}. Multi-modal 3D human pose
estimation and physical audiovisual commonsense reasoning address complementary
issues of modality imbalance, missing modalities, and implicit physical
relations \cite{qi2026balanced3dhpe,qi2026rdcl}. Our method similarly treats
visual, audio, depth, and layout cues as separate evidence sources, but it uses
them at test time through conservative answer gates rather than training a new
task-specific model.

Grounding and segmentation work is also relevant because many \vrrqa questions
depend on binding a queried object or person to the correct reference entity.
In-context image and video segmentation methods such as DC-SAM show how visual
prompts can transfer object-level masks across images and videos
\cite{qi2026dcsam}. In this report, we use the same principle at the reasoning
level: each ledger first makes the target, reference, temporal scope, and
coordinate frame explicit before any answer revision is allowed.

\section{Method}
\begin{figure*}[t]
\centering
\includegraphics[width=0.98\linewidth]{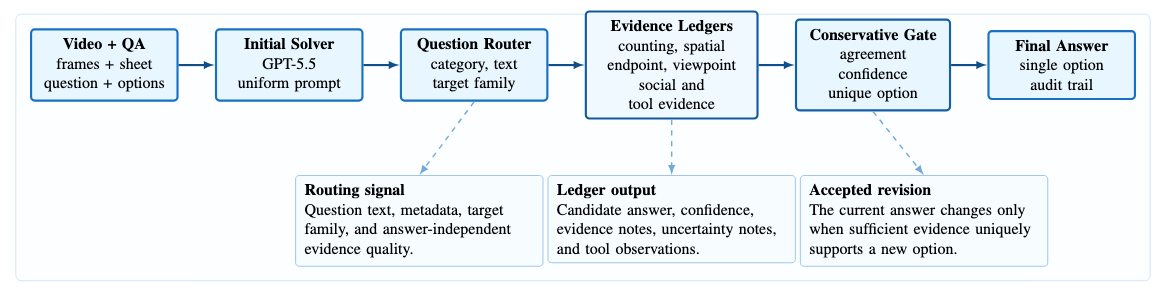}
\caption{Overview of the routed evidence-ledger pipeline. The initial solver
answers every question, the router invokes only relevant ledger families, and
the gate accepts a revision only when the candidate is supported by sufficient
evidence and maps uniquely to an option.}
\label{fig:pipeline}
\end{figure*}

Figure~\ref{fig:pipeline} gives an overview of the full test-time pipeline. The
system obtains a complete initial answer set, routes only the relevant question
families to evidence ledgers, and applies conservative gates to decide whether a
candidate should replace the current answer. Question interpretation is folded
into the solver and ledger prompts rather than implemented as a separate module:
each routed ledger is prompted to identify the target, operation, count unit,
and reference frame before mapping evidence to answer options.

\subsection{Initial Solver}
The first stage runs GPT-5.5 on every clip and question with a single prompt
profile. The video representation uses 64 sampled frames and a contact sheet
with four columns and 384-pixel cells. Frames are resized to a maximum side of
768 pixels. Decoding uses temperature 0, a 3000-token output budget, and high
reasoning effort through the API configuration. The model is asked to treat the
clip as a continuous event, audit all options, and return a JSON object
containing a legal answer choice and concise evidence.

This all-question stage is the only component that must answer from scratch. All
later components act as verifiers or candidate generators over routed subsets.

\subsection{Question and Metadata Routing}
The pipeline routes questions by broad task families. The router uses
only information available before scoring: question and option text, coarse
metadata, target family, and answer-independent visual or audio evidence
quality. Coarse category metadata is treated as a weak routing signal rather
than a source of truth; the question wording and visual evidence determine the
actual reasoning path.

The main routed families are \emph{Inferred Counting}, \emph{Vertical Spatial
Reasoning}, \emph{Lateral Spatial Reasoning}, \emph{Relative Depth and
Proximity}, \emph{Motion and Trajectory Dynamics}, and \emph{Viewpoint and
Visibility}. The router sends each family to prompts with different evidence
requirements. For example, counting experts are required to build a count ledger
before mapping to options, while spatial experts must explicitly bind the target
and reference and name the coordinate frame.

\subsection{Structured Evidence Ledgers}
Instead of asking a model to freely re-answer a difficult question, each routed
module produces a ledger. This follows the spirit of explicit reasoning traces
and reasoning-action decomposition \cite{wei2022cot,yao2023react}, but the
trace is constrained to answer-relevant visual evidence rather than free-form
deliberation. Each ledger is prompted to identify the target entities or events,
the requested operation, the count unit when applicable, the temporal or spatial
scope, and the reference frame for relational words. It then records evidence in
the corresponding family. Table~\ref{tab:ledgers} summarizes the ledger
families and the shared output schema. The fields are consumed by the gate
rather than treated as an automatic replacement.

\begin{table*}[t]
\centering
\caption{Question-aware ledger families and common output format.}
\label{tab:ledgers}
\small
\setlength{\tabcolsep}{4pt}
\renewcommand{\arraystretch}{1.08}
\begin{tabular}{@{}>{\raggedright\arraybackslash}p{0.14\textwidth}
                  >{\raggedright\arraybackslash}p{0.54\textwidth}
                  >{\raggedright\arraybackslash}p{0.27\textwidth}@{}}
\toprule
Ledger & Recorded evidence & Output fields \\
\midrule
Counting & Visible instances, distinct identities, completed events, repeated
views, and open-high options such as ``N or more''. &
Candidate answer, confidence, evidence notes, uncertainty notes. \\
Spatial & Target/reference binding, screen axis, intrinsic actor axis,
front/behind depth, vertical support, and same-height distractors. &
Candidate answer, confidence, coordinate frame, evidence notes. \\
Endpoint & Start state, final state, and whether the question explicitly asks
for the end of the clip. &
Candidate answer, confidence, start/end evidence, uncertainty notes. \\
Social/viewpoint & Speaker, addressee, reaction shot, line of sight, and
possible occlusion or device blocking. &
Candidate answer, confidence, grounding notes, tool observations. \\
\bottomrule
\end{tabular}
\end{table*}

\subsection{External Evidence}
External tools are used as evidence providers, not as direct answerers.
Open-vocabulary detection and cross-frame tracking, implemented with
GroundingDINO-style evidence \cite{liu2023groundingdino} and tracking cues
inspired by ByteTrack \cite{zhang2022bytetrack}, provide conservative lower
bounds for object counting. Depth cues, following the monocular depth
estimation setup of Depth Anything \cite{yang2024depthanything}, and
target-reference co-visibility help spatial verification. Pair crops focus
orientation verifiers on the relevant two entities. ASR, following the
Whisper-style weakly supervised speech recognition setup
\cite{radford2023whisper}, helps dialogue and speaker-addressee questions,
while layout scene-graph prompts organize multi-person and between-object
relations.

This distinction matters: a detector count may be incomplete, and an ASR segment
may not by itself identify the visual addressee. The system therefore only uses
tool outputs when they support a unique option mapping or agree with independent
model ledgers.

\subsection{Conservative Gates}
The merge rule defaults to keeping the current answer. A candidate may replace
the answer only when the stage-specific gate passes. Typical gates require
multi-expert agreement, evidence sufficiency, a minimum confidence threshold,
unique option mapping, or a detector lower bound that crosses an option
boundary. Counting gates often allow only upward corrections when the evidence
is a lower bound. Spatial gates require a clear coordinate frame and target
binding. Endpoint gates require evidence from the final window rather than a
salient middle frame.

The gate is category-risk aware. Validation experiments show that simply adding
another model ledger can create plausible but wrong agreement, especially for
lateral, depth, endpoint, and viewpoint questions. We therefore use additional
ledgers as diagnostic evidence unless the category-specific rule has shown high
precision. In the final evidence-gated pipeline, high-risk families require
stricter support such as target-reference grounding, depth or tracking
evidence, pair-crop verification, ASR/layout support, or independent
high-confidence ledgers before replacing the current answer.

\section{Implementation Details}
We do not train or fine-tune any model. All improvements come from test-time
prompting, routing, evidence extraction, and deterministic merging. The main
solver uses GPT-5.5 with temperature 0. Each video is represented by uniformly
sampled frames and a contact sheet so that the model can inspect both local
frames and the global temporal layout. The same visual preprocessing parameters
are used across stages unless a routed evidence module explicitly requires an
endpoint-focused window.

\begin{table}[t]
\centering
\caption{Implementation setup for the main solver.}
\label{tab:setup}
\begin{tabular}{lc}
\toprule
Component & Setting \\
\midrule
Base model & GPT-5.5 \\
Training & none \\
Frames per clip & 64 \\
Contact sheet & 4 columns, 384 px cells \\
Image max side & 768 px \\
Decoding temperature & 0 \\
Output format & JSON answer and evidence \\
\bottomrule
\end{tabular}
\end{table}

\section{Experiments}
\subsection{Main Pipeline}
Table~\ref{tab:main} summarizes the main pipeline. The initial solver provides
a strong all-question baseline. Broad category routing handles the largest
failure families with specialized prompts. Counting and spatial ledgers add
explicit instance, event, and coordinate-frame structure. Spatial evidence
verifiers introduce detection, depth, endpoint, and pair-crop signals. The final
pipeline adds target-family counting gates, person-reference consensus,
pairwise arbitration, ASR dialogue evidence, layout scene graphs, and
detector-track lower bounds.

The staged evaluation makes each component's contribution inspectable while
keeping the final pipeline deterministic. The code package also includes the
configurations used to regenerate candidate runs from the same prompt and
routing design.

\begin{table}[t]
\centering
\caption{Main pipeline ablation on the challenge test split.}
\label{tab:main}
\begin{tabular}{lcc}
\toprule
System & Test Acc. & Gain \\
\midrule
Initial GPT-5.5 solver & 71.89 & -- \\
+ question/metadata router & 74.92 & +3.03 \\
+ counting and spatial ledgers & 80.42 & +5.50 \\
+ spatial evidence verifier bundle & 86.40 & +5.98 \\
+ full evidence-gated pipeline & 92.95 & +6.55 \\
\bottomrule
\end{tabular}
\end{table}

Table~\ref{tab:splits} reports final performance on the challenge test split.

\begin{table}[t]
\centering
\caption{Final pipeline performance on the challenge test split.}
\label{tab:splits}
\begin{tabular}{lccc}
\toprule
Setting & Questions & Overall Acc. & Macro Acc. \\
\midrule
Test final pipeline & 172 & 92.95 & 93.79 \\
\bottomrule
\end{tabular}
\end{table}

The final pipeline obtains 92.95\% overall accuracy and 93.79\% macro accuracy
on the test split. Improvements
are largest on question families where the initial solver's failure modes are
systematic: counting, question-reference ambiguity, coordinate-frame reasoning,
endpoint orientation, person-reference relations, and dialogue/layout grounding.

\subsection{Merge-Gate Diagnostic}
We use validation as a diagnostic split for merge-rule design. To isolate the
effect of the gate, we run a controlled comparison on a fixed 200-question
validation subset that preserves the broad category mix of the full validation
split. All policies are evaluated on the same routed candidate pool, and only
the acceptance rule changes. This diagnostic therefore measures the precision
of accepted revisions rather than full-pipeline validation accuracy.

\begin{table}[t]
\centering
\caption{Merge-gate diagnostic on the fixed 200-question validation subset. All
rows use the same set of ledger-generated candidate answers; only the gate
policy changes.}
\label{tab:val200_gate}
\small
\setlength{\tabcolsep}{3.2pt}
\resizebox{\columnwidth}{!}{%
\begin{tabular}{@{}lrrrr@{}}
\toprule
Gate policy & Accepted & Fixes & Breaks & Prec. (\%) \\
\midrule
Raw two-ledger agreement & 6 & 2 & 4 & 33.3 \\
Conservative evidence gate & 2 & 2 & 0 & 100.0 \\
Three-source unlock gate & 5 & 1 & 4 & 20.0 \\
\bottomrule
\end{tabular}
}
\end{table}

Table~\ref{tab:val200_gate} shows why the final system does not use raw
agreement as a blanket replacement rule. Two-ledger agreement accepts more
changes, but most of them are breaks. The conservative evidence gate accepts
fewer revisions, but all accepted revisions are fixes in this subset. Adding a
third confirmation source is not automatically safer: the additional source can
reinforce the same wrong coordinate-frame interpretation and create
high-confidence but incorrect replacements. This supports the main design
choice of the final system: ledgers are valuable candidate generators, but
high-risk families need evidence-specific gates rather than generic voting.

\subsection{Design Choices}
The key design choice is to separate candidate generation from answer
acceptance. Broad prompt ensembling alone is not reliable: generic ``verify the
previous answer'' and free-form re-answer prompts can produce new correct
candidates, but without a strict gate they also break previously correct
answers. The useful component is not another full answer pass; it is the
combination of structured evidence and conservative acceptance.

We therefore evaluate modules through their routed coverage, accepted changes,
and aggregate accuracy changes. Question-reference disambiguation and broad
routing contribute the largest early gains. Counting and spatial ledgers improve
cases where the initial solver confuses instances, events, ordering references,
or coordinate frames. External evidence is most useful when it provides an
independent lower bound or a focused target-reference view. The conservative
gate is essential because it converts noisy candidates into high-precision
answer revisions.

\section{Implementation and Reproducibility}
The released code includes frame extraction, prompt profiles, routing utilities,
ledger-output parsing, and deterministic merge gates. We provide the saved
prediction files and configurations needed to replay the final submitted
predictions without additional model calls. The same codebase can also
regenerate candidate pools from the official videos and question files when API
credentials are available; because hosted model behavior may change over time,
the artifact replay path is the exact reproduction path for the submitted
prediction files. Runtime setup, credentials, generated frame caches, category
metadata generation, and command-level usage are documented in the README rather
than repeated here.

\section{Discussion}
The system is strongest on categories where failure modes are systematic:
counting, reference-frame disambiguation, coordinate-frame reasoning,
final-state orientation, person-reference depth/lateral confusion, and
dialogue/layout relations. It is less effective when the required evidence is
absent from sampled frames, when the question wording remains genuinely
ambiguous even after question-aware evidence checks, or when causal intent must
be inferred without explicit visual or audio support.

The central lesson is that high-performing video QA on \vrrqa is less about
asking a stronger model to guess again and more about forcing the model to
externalize the right evidence. Ledgers make the failure mode visible; tools
provide independent checks; conservative gates prevent evidence from becoming
overconfident answer replacement.

{\small
\bibliographystyle{plainnat}
\bibliography{references}
}

\end{document}